\documentclass[10pt,twocolumn,letterpaper]{article}

\usepackage{cvpr}              

\usepackage[accsupp]{axessibility}
\usepackage[dvipsnames]{xcolor}

\definecolor{cvprblue}{rgb}{0.21,0.49,0.74}
\usepackage[pagebackref,breaklinks,colorlinks,citecolor=cvprblue]{hyperref}

\def\paperID{70} 
\def\confName{CVPR}
\def\confYear{2024}

\title{Test-Time Adaptation with SaLIP: A Cascade of  SAM and CLIP for Zero-shot Medical Image Segmentation}

\author{
    Sidra Aleem\textsuperscript{1},
    Fangyijie Wang\textsuperscript{2},
    Mayug Maniparambil\textsuperscript{1},
    Eric Arazo\textsuperscript{3},
    Julia Dietlmeier\textsuperscript{4},\\
    Guenole Silvestre \textsuperscript{2},
    Kathleen Curran\textsuperscript{2},
    Noel E. O'Connor\textsuperscript{1},
    Suzanne Little\textsuperscript{4}
}

\begin{document}

\maketitle

\footnotetext[1]{ML-Labs, Dublin City University}
\footnotetext[2]{ML-Labs, University College Dublin}
\footnotetext[3]{Centre for Applied AI (CeADAR), University College Dublin, Ireland}
\footnotetext[4]{Insight SFI Centre for Data Analytics, Dublin City University}

\begingroup
\renewcommand\thefootnote{}\footnotetext{Corresponding author: \href{mailto:sidra.aleem2@mail.dcu.ie}{sidra.aleem2@mail.dcu.ie}}
\endgroup
\begin{abstract}
The Segment Anything Model (SAM) and CLIP are remarkable vision foundation models (VFMs). SAM, a prompt-driven segmentation model, excels in segmentation tasks across diverse domains, while CLIP is renowned for its zero-shot recognition capabilities. However, their unified potential has not yet been explored in medical image segmentation. To adapt SAM, to medical imaging, existing methods primarily rely on tuning strategies that require extensive data or prior prompts tailored to the specific task, making it particularly challenging when only a limited number of data samples are available. This work presents an in-depth exploration of integrating SAM and CLIP into a unified framework for medical image segmentation. Specifically, we propose a simple unified framework, SaLIP, for organ segmentation. Initially, SAM is used for part-based segmentation within the image, followed by CLIP  to retrieve the mask corresponding to the region of interest (ROI) from the pool of SAM's generated masks. Finally, SAM is prompted by the retrieved ROI to segment a specific organ. Thus, SaLIP is training/fine-tuning free and does not rely on domain expertise or labeled data for prompt engineering. Our method shows substantial enhancements in zero-shot segmentation, showcasing notable improvements in DICE scores across diverse segmentation tasks like brain (63.46\%), lung (50.11\%), and fetal head (30.82\%), when compared to un-prompted SAM. Code and text prompts are available at \href{https://github.com/aleemsidra/SaLIP}{SaLIP}.
\end{abstract}    
\section{Introduction}
\label{sec:intro}
The utilization of Vision-Foundation models (VFMs) has become increasingly prominent in various vision-related tasks, predominantly due to their zero-shot transfer capabilities to various downstream tasks. The Segment Anything Model (SAM)~\cite{kirillov2023segment} and Contrastive Language-Image Pre-Training (CLIP)~\cite{radford2021learning} have showcased remarkable generalization capabilities in segmentation and recognition, respectively. SAM, in particular, has been trained with a massive dataset of over 1 billion masks, making it highly adaptable to a wide range of downstream tasks through interactive prompts. SAM can be utilized to either segment everything in an image or to segment a specific region based on the prompts. SAM has shown impressive results in a broad range of tasks for natural images but its performance has been subpar when directly applied to medical images~\cite{mazurowski2023segment,deng2023segment,zhou2023can,hu2023sam}.  On the other hand, CLIP’s training with millions of text-image pairs has given it an unprecedented ability in zero-shot visual recognition.



Both SAM and CLIP have shown remarkable zero-shot transfer capabilities in various downstream tasks for natural images. However, their unified potential in the challenging domain of medical imaging has not yet been explored. 


While SAM  offers considerable advantages, there are inherent limitations to its application in medical image segmentation. SAM relies on prompts to segment specific regions. This prompt engineering requires domain expertise and manual intervention. However, it is particularly challenging in medical imaging due to the scarcity of high-quality labeled medical data and the need for specialized domain expertise. To address this, several studies have integrated SAM with other foundation models such as GroundingDINO~\cite{liu2023grounding} and YOLOV8~\cite{Jocher_Ultralytics_YOLO_2023} to generate bounding box prompts~\cite{pandey2023comprehensive}  for regions of interest (ROI)~\cite{biswas2023polyp}. 
These models are not directly applicable to medical image segmentation. To effectively utilize them for this purpose, they must undergo training with medical datasets containing images paired with their corresponding annotated masks. Their performance is also reliant on the size of training data, which requires careful evaluation and experimentation to achieve optimal segmentation results.
\begin{figure}[t]
  \centering
   \includegraphics[width=0.95\linewidth, height= 2cm]{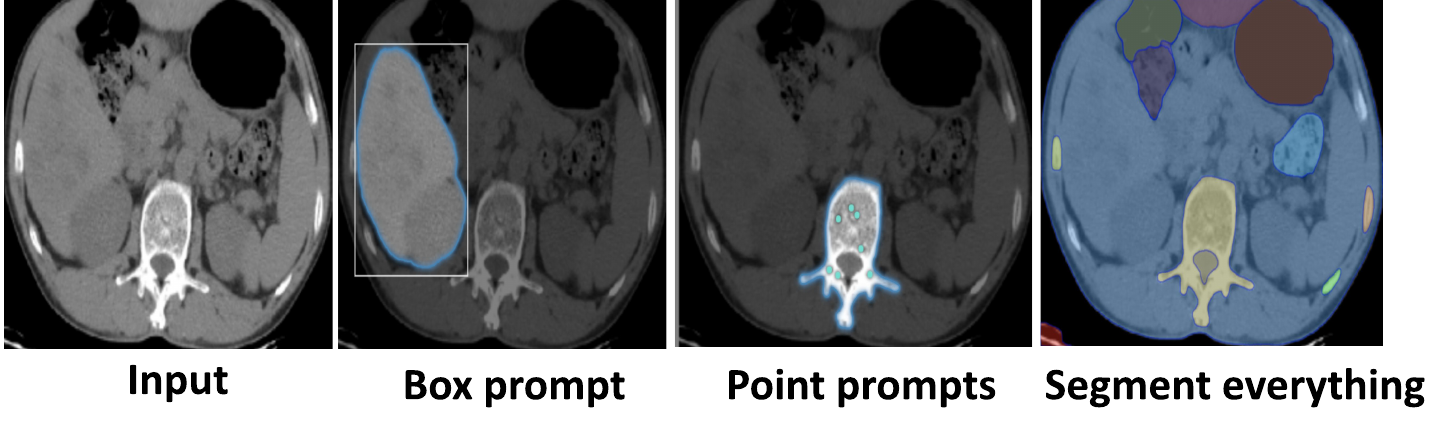}
   \caption{SAM efficiently segments regions based on prompts such as box or point prompts etc. However, such prompt generation needs domain expertise or annotated data, which is not readily available in medical imaging. To overcome this challenge, we use the segment everything mode to get the mask for every part in the image. Then, using CLIP, we select the mask corresponding to the specific organ and use it to generate prompts.}
   \label{fig:motivation}
\end{figure}

Furthermore, although SAM's capability of automatically segmenting everything in the image is appealing, there are further challenges to its application to medical imaging. One of the main challenges lies in the inherent variability of segmentation tasks. For example, given a liver cancer CT image, the segmentation task can vary depending on the specific clinical scenario. 
One clinician may be focused on segmenting the liver tumor, whereas another may require segmentation of the entire liver along with the surrounding organs. Additionally, clinicians are primarily interested in analyzing specific anatomical organs such as the liver, kidneys, spleen, lesions, etc. It becomes challenging to discern and focus on regions of interest amidst the growing number of segmented areas. Thus, such challenges impede the direct application of SAM to medical image segmentation.

To address the aforementioned challenges,  we leverage the combined capabilities of SAM and CLIP and introduce a unified framework called SaLIP, for zero-shot organ segmentation. SAM effectively performs organ segmentation with prompts as shown in ~\cref{fig:motivation}, but its effectiveness hinges on domain expertise and annotated data for prompt engineering, which is not readily available in the medical domain. To circumvent these challenges, we adopt the segment everything mode to segment every part in the image and cascade it with CLIP to get the mask for specific organs. 

Initially, our framework SaLIP, employs SAM to automatically segment every part within the image. SAM provides exhaustive segmentation, however the resulting masks lack semantic labels. To extract the relevant ROI mask from the pool of generated masks, we first crop the original image according to these masks. This set of cropped regions is passed to CLIP. By employing visually descriptive (VDT) sentences related to the target organ, CLIP then retrieves the corresponding crop in a zero-shot manner~\cite{maniparambil2023enhancing}. The VDT prompts for CLIP are generated via GPT-3.5 ~\cite{achiam2023gpt}. Finally, the retrieved ROI mask is used for bounding box prompt generation, which is eventually used to prompt SAM to guide the specific organ segmentation. Hence, our framework is training/fine-tuning free and independent of domain expertise or labeled data for prompt engineering. By combining the strengths of SAM and CLIP, our method effectively performs zero-shot medical organ segmentation. We conduct experiments across three diverse medical imaging datasets encompassing MRI scans, ultrasound, and X-ray images to demonstrate the effectiveness of SaLIP.

Our contributions can be summarized as follows:
\begin{itemize}
    \item We propose a simple unified framework that leverages the combined capabilities of SAM and CLIP for medical image segmentation. We demonstrate that a cascade of these foundation models can improve the zero-shot segmentation accuracy in medical imaging. 

    \item To effectively address the challenges associated with applying SAM directly to medical imaging and to optimize its utilization for medical image segmentation, we propose employing both segment everything and promptable segmentation modes. To the best of our knowledge, we are the first to investigate the utilization of SAM's dual modes for zero-shot medical imaging segmentation.

    \item Our unified framework SaLIP is adapted fully at test-time for zero-shot medical image segmentation, thereby efficiently alleviating the training costs associated with these foundation models. By leveraging Large language models (LLMs), our method eliminates the need for domain expertise in prompt engineering.

\end{itemize}

\section{Related Work}
\subsection{Segment Anything Model (SAM)}
SAM~\cite{kirillov2023segment} is a promptable vision foundation segmentation model that aims to segment everything in an image conditioned on different kinds of prompts like bounding boxes and point prompts. It presents a new data engine and portable model for general object segmentation. Given prompts, SAM returns valid segmentation masks. It has three modules: an image encoder, a prompt encoder, and a mask decoder. Masked Autoencoders (MAE)~\cite{he2022masked}, a pre-trained Vision Transformer (ViT)~\cite{dosovitskiy2020image} is used as an image encoder.  The mask decoder efficiently maps the image embedding, prompt embedding, and an output token to a mask.

\begin{figure*}[t]
  \centering

\includegraphics[width=0.9\linewidth]{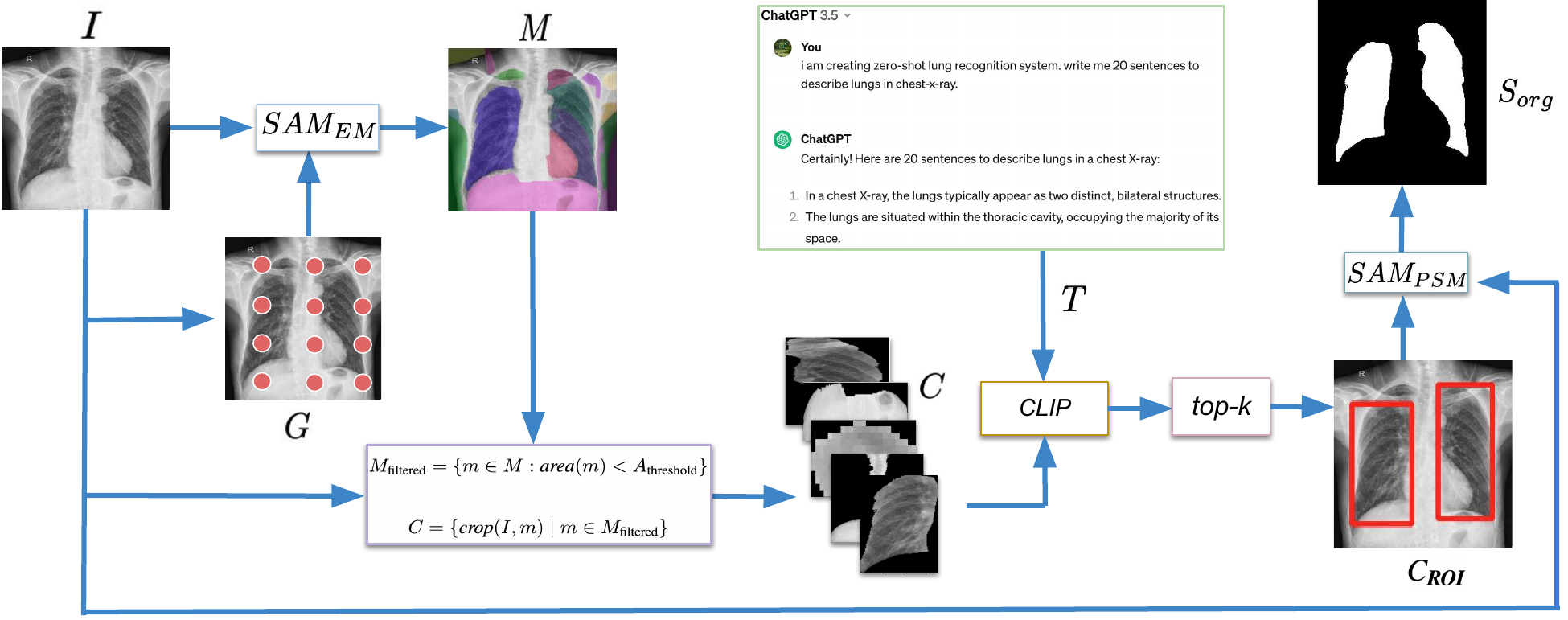}
\caption{Illustration of SaLIP: $\mathbf{\textit{SAM}}_{\mathbf{\textit{EM}}}$ segments the input image \textit{\textbf{I}} using a grid-wise set of keypoints $\textit{\textbf{G}}$, as prompts to produce part-based segmentation masks \textit{\textbf{M}}. To remove $\mathbf{\textit{m}} \in \mathbf{\textit{M}}$ that corresponds to background, a {$\mathbf{\textit{A}}_{\mathbf{\textit{\textbf{threshold}}}}$} is applied on \textit{\textbf{M}}. Subsequently, \textit{\textbf{I}} is cropped based on $\mathbf{\textit{M}}_{\textit{\textbf{filtered}}}$ to generate a set of crops \textit{\textbf{C}}, which are then fed into CLIP along with visually descriptive sentences \textit{\textbf{T}} generated by GPT 3.5. The region of interest crops $\mathbf{\textit{C}}_{\textbf{\textit{ROI}}}$ are retrieved from CLIP using $\textbf{argmax}$, and $\textbf{top-k}$ crops (two in this case) are selected as ROI. The extracted ROI is leveraged to generate bounding box prompts (coordinates shown in red), which are used to prompt $\mathbf{\textit{SAM}}_{\textbf{\textit{PSM}}}$ to get the specific organ segmentation $\textbf{\textit{S}}_{\textbf{\textit{org}}}$ within \textit{\textbf{I}}.}  
\label{fig:framework}
\end{figure*}

\subsection{SAM for medical image segmentation}
The application of SAM has been investigated within the medical domain. The first line of research focuses on the adaptability of SAM using fine-tuning strategies. MedSAM~\cite{ma2024segment} fine-tunes the SAM mask decoder on large-scale datasets, SAMed~\cite{zhang2023customized} adopts a low-rank-based fine-tuning strategy (LoRA)~\cite{hu2021lora}, and trains a default prompt for all images in the dataset. Medical SAM Adapter (MSA)~\cite{wu2023medical} uses adapter modules for fine-tuning.  These approaches yield promising results, often matching or surpassing state-of-the-art fully-supervised models. Nevertheless, these SAM-based methodologies still require substantial amounts of data for supervised fine-tuning and do not fully leverage the prompting ability.
The second line of research focuses on evaluating the performance of few-shot segmentation by prompting SAM to return a specific object segmentation. Several recent studies, including~\cite{deng2023segment,huang2024segment,hu2023sam,mattjie2023exploring,mohapatra2023brain} have evaluated SAM’s capability on different medical image segmentation tasks in the context of zero-shot transfer.  However, this prompt generation requires domain expertise or high-quality labeled data. 

In contrast, our method is entirely independent of training or domain expertise for prompt engineering. Instead, it effectively adapts SAM to medical imaging segmentation by harnessing the capabilities of both segment-everything and promptable modes, using CLIP as the bridge between the two. Our framework facilitates fully test-time zero-shot organ segmentation in medical imaging.

\subsection{CLIP}
CLIP~\cite{radford2021learning} is a pre-trained large Vision-Language Model (VLM) known for its strong generalizability and impressive zero-shot domain adaption capabilities. An effective method for adapting CLIP to various domains is through prompt engineering, a process that typically incorporates relevant semantic details related to the specific target task~\cite{maniparambil2023enhancing}. CLIPSeg~\cite{luddecke2022image} extends the CLIP model with a transformer-based decoder that facilitates dense prediction. MedCLIP~\cite{Wang2022medclip} fine-tunes the CLIP model by separating medical images and texts to expand the available training data exponentially at a low cost. CXR-CLIP~\cite{you2023cxrclip} improves its performance in chest X-ray classification tasks by fine-tuning the CLIP image and text encoders using samples from image-text and image-label datasets. These methodologies require supervised fine-tuning on medical image-text pairs. Other studies such as~\cite{liu2023clip,zhang2023medical,zhang2023bmedical} have demonstrated that the incorporation of text embeddings learned from CLIP into medical segmentation models achieves state-of-the-art results. However, these medical image-text pairs are collected under guidelines and with the support of domain experts.

\section{Methodology}
\label{sec:methodology}
In this section, we first review  SAM and CLIP in \cref{sec:SAM} and ~\cref{clip}. Subsequently, we explain our unified framework SaLIP in ~\cref{sec:methodolgy}.  Our framework is illustrated in ~\cref{fig:framework}.

\subsection{Preliminaries}
\label{sec:preliminaries}
\subsubsection{SAM}
\label{sec:SAM}
SAM is a prompt-driven segmentation foundation model. It consists of three main components: an image encoder, a prompt encoder, and a lightweight mask decoder. 
We denote an input image as $\textbf{\textit{I}} \in \mathbb{R}^{H \times W \times 3}$ and an input visual prompt as $\textbf{\textit{P}} \in \mathbb{R}^{N}$, where $\textbf{\textit{$\textbf{H} \times \textbf{W}  $}}$  are the spatial dimensions and  \textit{\textbf{N}} is the number of prompts. The image encoder is a MAE~\cite{he2022masked} pre-trained Vision Transformer (ViT)~\cite{dosovitskiy2020image}. It encodes an image into dense features $\mathbf{\textit{F}}_{\mathbf{\textit{SAM}}} \in \mathbb{R}^{\frac{H}{16} \times \frac{Q}{16}} $. The prompt encoder encodes prompts \textit{\textbf{P}} into sparse prompts $\mathbf{\textit{Q}}_{\mathbf{\textit{sp}}}$. \textit{\textbf{P}} can either be sparse, such as points, boxes, or text, or dense, like masks. The points and boxes are represented by positional encodings ~\cite{tancik2020fourier} summed with learned embeddings for each prompt type. Currently, SAM  does not directly process text prompts and the text-to-mask task is still in its exploratory stages and is not entirely robust~\cite{kirillov2023segment}.

The mask decoder efficiently maps the image features $\mathbf{\textit{F}}_{\mathbf{\textit{SAM}}}$, $\mathbf{\textit{Q}}_{\mathbf{\textit{sp}}}$  , and an output token to a mask.  It uses prompt self-attention and cross-attention in two directions (prompt-to-image embedding and vice-versa) to update all embeddings. After running two blocks, a multilayer perceptron (MLP) maps the output token to a dynamic linear classifier, which then computes the mask foreground probability at each image location.

SAM can operate in two distinct modes: segment everything mode ($\mathbf{\textit{SAM}}_{\textbf{\textit{EM}}}$) and promptable segmentation mode $\mathbf{\textit{SAM}}_{\textbf{\textit{PSM}}}$.  The former can segment everything in the image without relying on externally provided prompts. Instead, a grid of keypoints is generated on the image, and these grid-wise set of keypoints are used as prompts to segment everything in the image. The latter segment a specific set of ROIs based on the prompts given to SAM. 

Our framework utilizes both modes of SAM with CLIP as a bridge between them.


\subsubsection{CLIP}
\label{clip}

Using contrastive pre-training on large image-text datasets, CLIP performs image classification. CLIP aligns image and text modalities within a shared embedding space. After pretraining, CLIP directly performs image classification on the target dataset without any fine-tuning. For an image {$\mathbf{\textit{I}} \in \mathbb{R}^{\mathbf{\textit{H}} \times \mathbf{\textit{W}} \times \mathbf{\textit{C}}}$}, where $\textbf{\textit{$\textbf{H} \times \textbf{W} \times \textbf{C} $}}$  denotes spatial dimension, the vision encoder  \textit{\textbf{f}} maps \textit{\textbf{I}} into a joint embedding space to get the image features {$\mathbf{\textit{E}} \in \mathbf{\textit{D}}$}
 with dimension \textit{\textbf{D}}. During inference, a prompt template such as \textit{`A photo of {classname}'} is used to generate sentences for $K$ different classes and passed through the text-encoder to yield classifier weight matrix \textbf{\textit{ $W \in \mathbb{R}^{D \times K}$ }}. Prediction probabilities are then calculated by multiplying image feature \textit{\textbf{f}} and \textit{\textbf{W}}  and applying a softmax function.

 In this work, to construct textual prompts for CLIP, we use ensembles of visually descriptive (VDT) information for each class~\cite{maniparambil2023enhancing}.

\subsubsection{SaLIP}
\label{sec:methodolgy}

In this section, we provide a detailed overview of our unified framework for zero-shot organ segmentation. First, we explain how $\mathbf{\textit{SAM}}_{\textbf{\textit{EM}}}$ generates masks for every part within the image, followed by an explanation of how CLIP retrieves the relevant ROIs from the pool of generated part-based masks. Finally, we illustrate how we leverage retrieved ROIs to create prompts for $\mathbf{\textit{SAM}}_{\textbf{\textit{PSM}}}$.

To generate part-based segmentation masks from the image, we use $\mathbf{\textit{SAM}}_{\textbf{\textit{EM}}}$. It generates an extremely exhaustive prediction of nearly any object or part in the images. It takes a grid-wise set of keypoints $\mathbb{\textbf{\textit{G}}} \in \mathbb{R}^{\textbf{\textit{g}}^{2} \times 2}$
 as input, where \textbf{\textit{g}} is the point number along one side of the image. Then the masks are generated by  $\mathbf{\textit{SAM}}_{\textbf{\textit{EM}}}$  by prompting it with a set of grid-wise key points.
\begin{equation}
    \mathbf{M} = \mathbf{SAM}_{\mathbf{EM}}(\mathbf{I}, \mathbf{G})
\end{equation}
where $\mathbf{\textit{I}} \in \mathbb{R}^{3 \times H \times W}$ is the input image, and {$\mathbf{\textit{M}} \in \mathbb{R}^{N \times H \times W}$}  is the set of all the part-based generated masks. $\textit{\textbf{N}}$  refers to the number of masks and $\textbf{\textit{$\textbf{H} \times \textbf{W}$}}$ is the spatial dimension. 

The process of generating part-based masks with $\mathbf{\textit{SAM}}_{\textbf{\textit{EM}}}$ is greatly influenced by the selection of hyperparameters utilized for the SAM mask generator module. 
 To streamline and achieve optimal part-based segmentation, we use a random search for optimal hyper-parameters for $\mathbf{\textit{SAM}}_{\textbf{\textit{EM}}}$ using five randomly selected images. The combination of hyper-parameters that yields the highest DICE score is used as the final configuration of $\mathbf{\textit{SAM}}_{\textbf{\textit{EM}}}$ to generate part-based masks for the entire dataset.

Following this, the next step in the pipeline is to extract the ROI mask from \textbf{\textit{M}} using CLIP. To accomplish this, we first utilize \textbf{\textit{M}} to crop \textbf{\textit{I}}, thereby producing a series of crops, each corresponding to a mask in \textbf{\textit{M}}. As $\mathbf{\textit{SAM}}_{\textbf{\textit{EM}}}$ generates masks for every element within the image, there arises the possibility of predicting a mask corresponding to the background. In such instances, the resulting crop has spatial dimensions identical to \textbf{\textit{I}}. Consequently, when these crops are subsequently passed to CLIP, there is a risk of miss-classification, as CLIP may perceive them as ROI due to the presence of relevant region in \textbf{\textit{I}} as discussed in ~\cref{area}. To mitigate this issue, instead of directly forwarding the entire set of masks \textbf{\textit{M}} to CLIP, we first filter out the masks {$\mathbf{\textit{m}} \in \mathbf{\textit{M}}$} that potentially correspond to the background region using area-based filtering on each mask within \textbf{\textit{M}}. To determine the optimal threshold for area-based filtering, we perform a random hyperparameter search within the space defined by the areas of masks in \textbf{\textit{M}}. This search is carried out simultaneously with the hyper-parameter optimization process for  $\mathbf{\textit{SAM}}_{\textbf{\textit{EM}}}$, using the same methodology discussed above. The area filtering is conducted, and set crops are generated as follows:
%

%
\begin{equation}
\mathbf{M_{\text{filtered}} = \{ m \in M : \textit{area}(m) < A_{\text{threshold}} \}}
\end{equation}
\begin{equation}
\mathbf{C} = \{ \mathbf{\textit{crop}}(\mathbf{I}, \mathbf{m}) \mid \mathbf{m} \in \mathbf{M}_{\mathbf{\text{filtered}}} \},
\end{equation}

where {$\mathbf{\textit{A}}_{\mathbf{\textit{threshold}}}$} is the value of area achieved via hyper-parameter search used for filtering \textbf{\textit{M}} based on area,
\textbf{\textit{m}} is a mask from \textbf{\textit{M}}, $\mathbf{\textit{M}}_{\textit{filtered}}$ is the set of masks after removing the \textbf{\textit{m}} corresponding to background. $\mathbf{\textit{crop}}(\mathbf{I}, \mathbf{m})$ denotes the operation of cropping \textbf{\textit{I}} according to {$\mathbf{\textit{m}} \in \mathbf{\textit{M}}_{\textit{filtered}}$}
 The set of generated crops \textbf{\textit{C}} along with the textual prompts are passed to CLIP to select the crop corresponding to ROI. 
 To construct textual prompts for CLIP, we use prompt ensembling, a technique that constructs several sentences for each class and subsequently averages the classification vectors. We use prompt ensembles of visually descriptive (VDT) information for each class ~\cite{maniparambil2023enhancing}. The VDT sentences are generated via GPT 3.5 and passed through CLIP to get the text embeddings and averaged to obtain a single text prototype $\mathbf{\textit{W}}_{\textbf{\textit{T}}}$ for the organ under consideration. Now all the image crops in \textbf{C} are passed through CLIP's vision encoder to obtain vision embeddings \(\textbf{\textit{E}}_{c}\). Subsequently, the mask corresponding to ROI is computed as:
 
%
\begin{equation}
\mathbf{C_{ROI} = \text{topk}\left(\arg\max_{c \in C} S(\textbf{\textit{E}}_{c}, \textbf{\textit{W}}_{\textit{T}})\right)}
\end{equation}

where {$\mathbf{\textit{S}}(\mathbf{\textit{E}}_{c},\mathbf{\textit{W}}_{T})$} represents a similarity function which computes cosine similarity between any embeddings \(\textbf{\textit{E}}_{c}\) of any crop $\textbf{\textit{c}} \in \textbf{\textit{C}}$ and the text embeddings \(\textbf{\textit{W}}_{\textbf{\textit{T}}}\). $\textbf{\textit{k}}$ denotes the number of ROIs and varies depending on the number of ROIs in the image. $\mathbf{\textit{C}}_{\textbf{\textit{ROI}}}$ is the mask corresponding to ROI.


Finally, we compute the bounding box prompts using the minimum and maximum $\textbf{\textit{X}}$, $\textbf{\textit{Y}}$ co-ordinates of the retrieved $\mathbf{\textit{C}}_{\textbf{\textit{ROI}}}$ and use it to prompt  $\mathbf{\textit{SAM}}_{\textbf{\textit{PSM}}}$ as:
{\begin{equation}
    \mathbf{S}_{\mathbf{org}} = \mathbf{SAM}_{\mathbf{PSM}}(I, P)
\end{equation}}
where $P \in \mathbb{R}^{k \times 4}$ is the bounding box computed from $\mathbf{\textit{C}}_{\textbf{\textit{ROI}}}$, $\textbf{\textit{N}}$ is the number of box prompts which varies according to ROI and $\textbf{\textit{S}}_{\textbf{\textit{org}}}$ is the final segmentation for ROI. 

\section{Experiments}
\label{sec:experiments}
\subsection{Datasets and Metrics}

We assessed our method across three diverse medical imaging modalities, encompassing two datasets focusing on single-organ segmentation and one more challenging dataset requiring the segmentation of two distinct organs.  Calgary-Campinas (CC359)~\cite{souza2018open} is a multi-vendor (GE, Philips, Siemens), multifield strength (1.5, 3) magnetic resonance (MR) T1-weighted volumetric brain imaging dataset. It has six different domains and contains 359 3D brain MR image volumes, primarily focused on the task of skull stripping. The HC18 \cite{Heuvel2018us} consists of 2D fetal head ultrasound images obtained throughout all trimesters of pregnancy. These images have been annotated with biometrics by experienced medical experts. From this dataset, a subset of 200 images is selected for testing purposes. X-ray Masks and Labels ~\cite{pandey2019} consists of 800 2D chest X-ray images, each accompanied by its corresponding mask for lung segmentation.  We use the DICE score (DSC) and mean intersection over union (mIoU) as our evaluation metrics.

\subsection{Implementation Details}
\label{implementation detail}
We employed  ViT-H, a variant of SAM, and ViT-L/14 trained in CLIP by OpenAI. For CLIP, the visually descriptive textual sentences are generated using GPT 3.5, the details can be found in the supplementary material,  ~\cref{sec:prompts}. We implemented our framework in PyTorch ~\cite{paszke2019pytorch} with SAM codebase \footnote{\url{https://github.com/facebookresearch/segment-anything}}. All experiments are performed on a desktop computer with the Ubuntu operating system 20.04.6 LTS with CUDA 11.6, NVIDIA GeForce RTX 3090 GPU. For reproducibility, a random seed is set to 1234. 

\begin{table*}[t]
  \centering
  \begin{tabular}{@{}llcccccccc@{}}
    \toprule
    ROI & Dataset & \multicolumn{2}{c}{U-Net} & \multicolumn{2}{c}{GT-SAM}  & \multicolumn{2}{c}{Un-prompted SAM} &\multicolumn{2}{c}{Ours}\\
     \cmidrule(lr){3-4} \cmidrule(lr){5-6} \cmidrule(lr){7-8} \cmidrule(lr){9-10}
     &  &DSC  & mIoU & DSC & mIoU  & DSC & mIoU & DSC & mIoU\\
    \midrule
    &GE 1.5 & 0.98 & 0.93 &   0.95 &  0.91 & 0.33 &  0.29 & \textbf{0.92}& \textbf{0.87}\\
    &Philips 1.5& 0.97 & 0.95 & 0.96  &  0.93 & 0.41 & 0.31 & \textbf{0.94}  & \textbf{0.85}\\
    Brain &Philips 3 & 0.95 & 0.92 &  0.93 & 0.89 & 0.40 & 0.39 & \textbf{0.89}&  0.80 \\
    &Siemens 1.5 & 0.97 &  0.95 &  0.95 & 0.91 & 0.39  & 0.26 & \textbf{0.90}  & \textbf{0.81}\\
    &Siemens 3 & 0.98 & 0.92 &  0.96 &  0.90 & 0.41 & 0.32 & \textbf{0.93} & \textbf{0.85}\\
    \midrule
    Lungs & X-ray & 0.98 & 0.95 & 0.94  & 0.90 & 0.47  & 0.31 & \textbf{0.83 }& \textbf{0.76} \\
    \midrule
    Fetal head & Ultrasound & 0.95 & 0.91 & 0.95 & 0.91 & 0.55 & 0.40 & \textbf{0.81} & \textbf{0.72}\\   
    \bottomrule
  \end{tabular}
  \caption{Comparison of our method with other baselines. Our method significantly outperforms un-prompted SAM, without using domain expertise or annotated data for prompt engineering. \textbf{Note:} GT-SAM uses the prompts extracted from ground truth.}
  \label{tab:tab1}
\end{table*}

\subsection{Results and Analysis}
We evaluated our proposed method against U-Net~\cite{ronneberger2015unet}, ground truth-aided SAM (GT-SAM), and un-prompted SAM. U-Net is widely used in medical image segmentation, and it is fine-tuned for all three of our datasets separately. GT-SAM is an upper bound in which box prompts for SAM are directly derived from the ground truth. As our method does not utilize ground truth for prompt generation, to ensure a fair comparison and simulate the real-world medical imaging scenarios without annotated data, we employ an un-prompted version of SAM. In this version, SAM is not provided with prompts from ground truth, rather it utilizes its default prompt embedding.

\begin{figure}[t]
  \centering
   \includegraphics[width=0.8\linewidth]{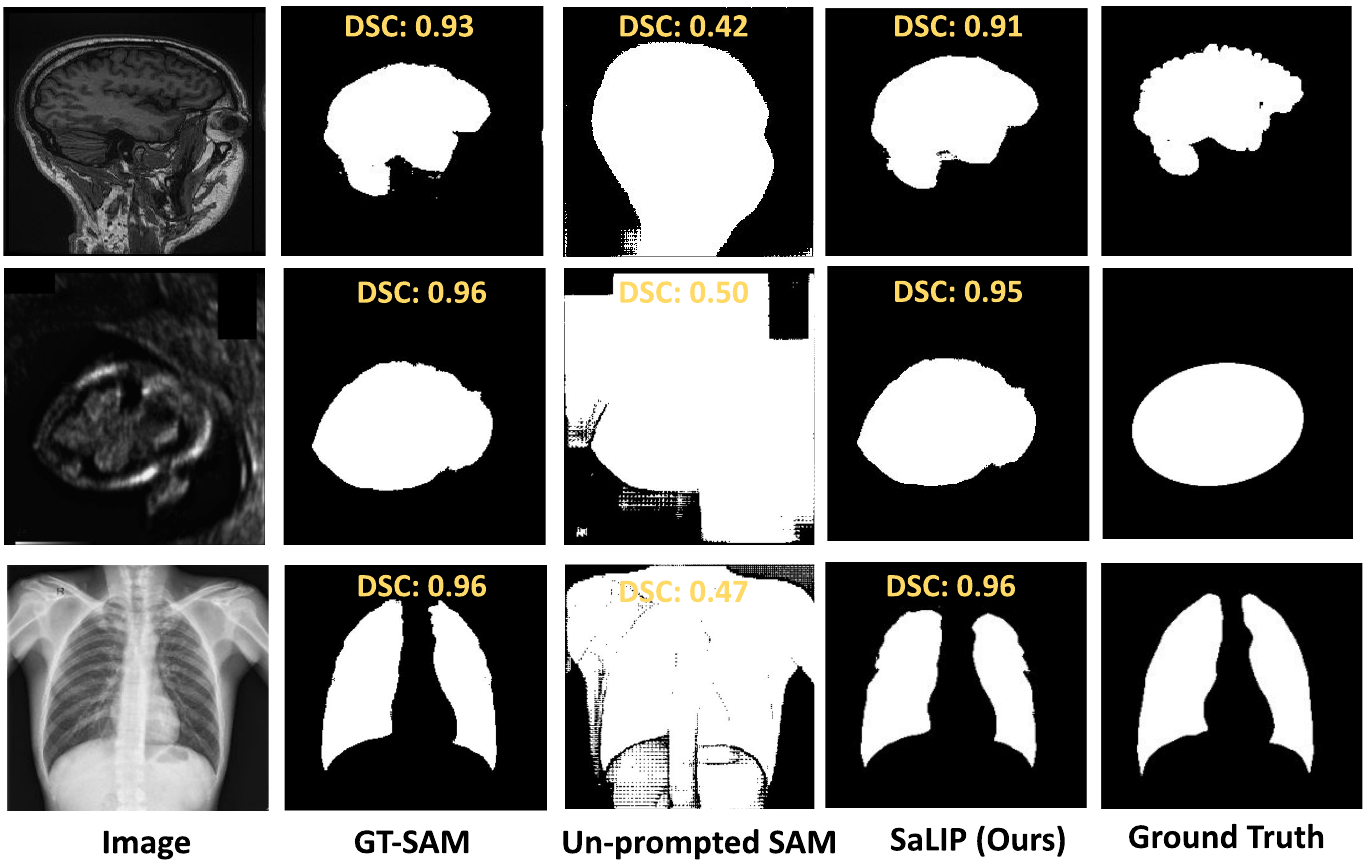}

   \caption{GT-SAM is the upper bound, un-prompted SAM, SaLIP (ours). The text in yellow refers to the DSC with each method respectively.}
   \label{fig:qn_results}
\end{figure}

The quantitative results are shown in ~\cref{tab:tab1}. The difference in performance among all methods can be attributed to the prompts utilized for SAM, highlighting the significant dependency of SAM's performance on the prompts employed. For CC359 ~\cite{souza2018open}, our method achieves an average of 0.94 DSC, significantly outperforming the un-prompted SAM's average DSC of 0.31. When evaluated for lung segmentation, our approach elevates the performance from an initial DSC of 0.31 with unprompted SAM to 0.83. Our method achieves an average DSC of 0.81 for segmenting the fetal head on HC18 ~\cite{Heuvel2018us}, achieving a 26\% increase compared to the unprompted SAM's average DSC of 0.55. The low performance of the un-promoted SAM is because it does not utilize any prompts, which leads to its inability to segment the ROIs. This demonstrates that SAM's ability to effectively segment is strongly reliant on the prompts. Furthermore, the qualitative analysis presented in ~\cref{fig:qn_results}, illustrates that the un-prompted SAM fails to accurately perform organ segmentation, as it generates a general segmentation mask for the regions in the image rather than delineating specific organs. Hence, SAM's applicability to medical imaging scenarios is limited, where obtaining domain expertise and annotated data for prompt engineering is challenging. On the other hand, GT-SAM achieves high DSC of 0.95, 0.94, and 0.91 for brain, lungs, and fetal head segmentation respectively. This high performance is attributed to GT-SAM's use of perfect prompts extracted directly from the ground truth. In contrast, our method performs fully test-time zero-shot organ segmentation without relying on external prompts or domain expertise, demonstrating its effectiveness and versatility. To evaluate our method in comparison to GT-SAM, it is important to highlight that GT-SAM benefits from perfect prompts extracted directly from ground truth and is the upper bound. In contrast, our method operates in a zero-shot manner completely independent of ground truth or any domain expertise for prompt engineering. Despite this, our method still achieves results comparable to GT-SAM. This demonstrates the effectiveness and adaptability of our approach in the context of organ segmentation in medical images, where access to domain expertise and perfect annotated data is either limited or impractical. To further demonstrate the effectiveness of our approach, 
the qualitative results are shown in ~\cref{fig:qn_results}. 

\begin{figure}[t]
  \centering
   \includegraphics[width=0.95\linewidth]{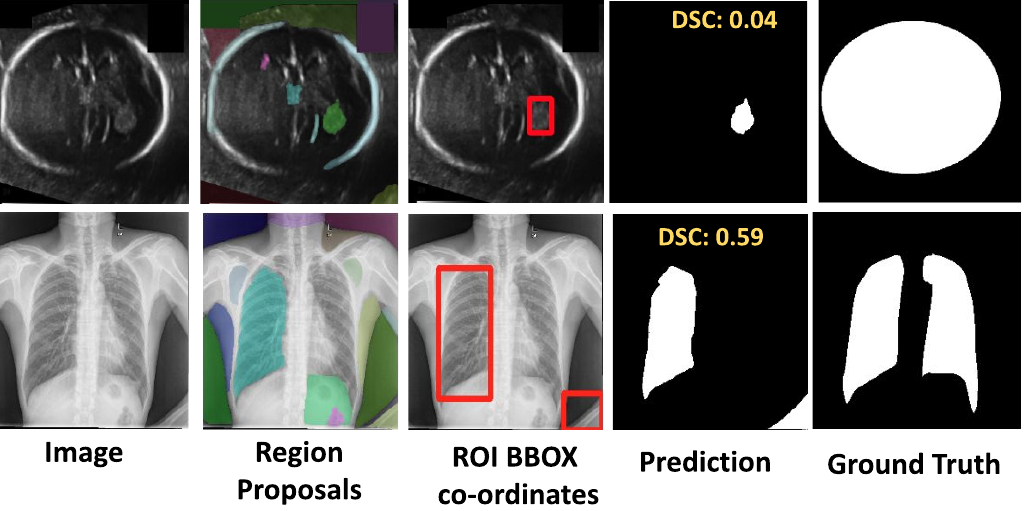}
   \caption{SAM failure cases: First row: $\mathbf{\textit{SAM}}_{\textbf{\textit{EM}}}$ fails to generate a mask for the fetal head, resulting in miss-classification by CLIP.
Second row: $\mathbf{\textit{SAM}}_{\textbf{\textit{EM}}}$ generates a mask for the right lung but fails to generate one for the left lung, leading to CLIP retrieving the wrong crop. }
   \label{fig:sam_limitation}
\end{figure}
\subsubsection{ Failure Cases and Future Work}
 Although our method demonstrates effective performance, through an in-depth analysis and exploration, we have identified two sets of limitations: one at the SAM level and the other one at the CLIP level. 
 
\textbf{\textit{SAM part-based segmentation:}} This refers to the instances where $\mathbf{\textit{SAM}}_{\textbf{\textit{EM}}}$ fails to generate a mask for the ROI as shown in ~\cref{fig:sam_limitation}. Among our three datasets, this issue is particularly prominent in ultrasound and X-ray images. Due to the nature of how ultrasound images are captured, inherent limitations in fetal ultrasound images are very common, such as acoustic shadows, speckle noise, and obscured boundaries. These characteristics pose challenges to the accurate generation of masks for the fetal head by the $\mathbf{\textit{SAM}}_{\textbf{\textit{EM}}}$. These issues often arise from sub-optimal selection of hyperparameters for SAM's mask generation process. To tackle this, we have implemented an automated hyperparameter search for $\mathbf{\textit{SAM}}_{\textbf{\textit{EM}}}$ hyperparameters. This automation significantly mitigated the problem.

\begin{figure}[t]
\centering
\includegraphics[width=0.9\linewidth, height = 2.7cm]{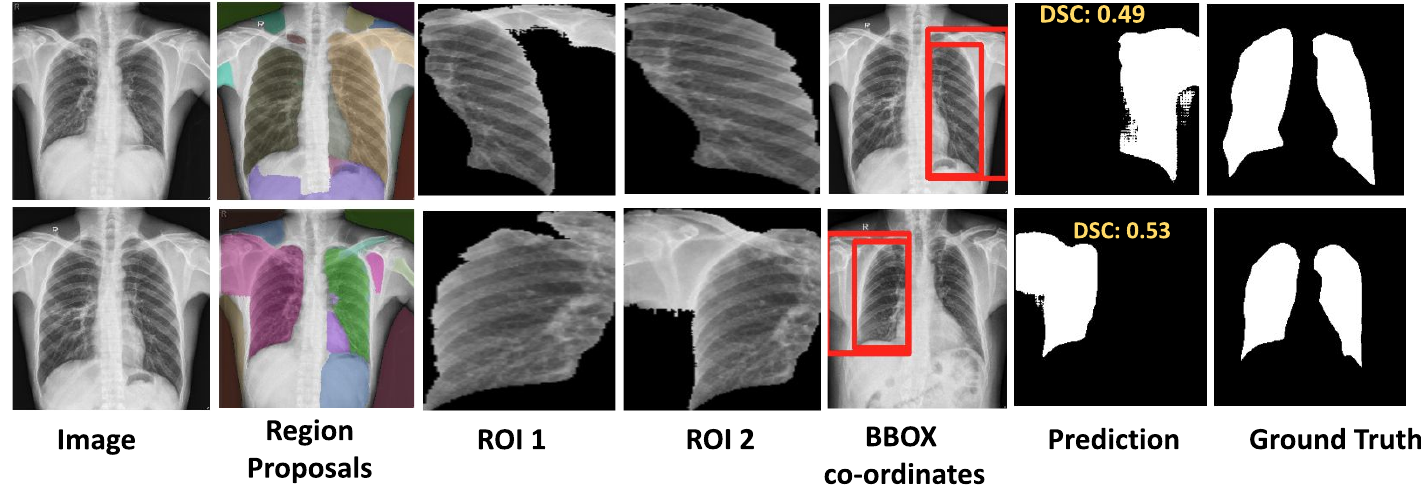}
\caption{CLIP failure cases: $\mathbf{\textit{SAM}}_{\textbf{\textit{EM}}}$ generates multiple masks for both ROIs (left and right lung). First row: CLIP while correctly recognizing the left lung, identifies a second mask for the same lung region and fails to retrieve the one for the right lung. Second row: CLIP does not retrieve the left lung crop.}
\label{fig:clip_limitation}
\end{figure}

\textbf{\textit{CLIP mask retrieval:}} There are instances where $\mathbf{\textit{SAM}}_{\textbf{\textit{EM}}}$ generates masks corresponding to ROIs, but CLIP fails to retrieve them. Such issues arise due to the generation of multiple masks for a single region by $\mathbf{\textit{SAM}}_{\textbf{\textit{EM}}}$ and impact the datasets with multiple ROIs, lungs in our case. CLIP in some cases retrieves the masks corresponding to the same lung region as shown in ~\cref{fig:clip_limitation}. 

For extracting lung crops from the pool of masks generated by $\mathbf{\textit{SAM}}_{\textbf{\textit{EM}}}$ via CLIP, we use a single set of visually descriptive prompts that characterize both lungs in a chest X-ray (supplementary ~\cref{sec:lungs}). To address the challenge of the same region mask retrieval by CLIP, we experimented with a separate set of prompts for the left and right lungs and evaluated the impact. For more details, please refer to the supplementary material (\cref{sec:separate_prompts}). However, our results demonstrated that CLIP shows limited performance in precise localization and recognition tasks and lacks semantic knowledge in distinguishing objects based on their spatial alignment (i.e., left and right). Consequently, employing separate sets of prompts to describe organs based on their spatial alignment does not mitigate the issue. The results are presented in \cref{tab:prompts}. In contrast to using separate prompts, our approach of utilizing a single set of prompts describing both lungs achieves 0.83 DSC, thereby outperforming the separate prompts, which achieve 0.67 and 0.28 DSC for the left and right lung, respectively.

\begin{table}[!hb]
  \centering
  \begin{tabular}{@{}lccc@{}}
    \toprule
    & Right Lung & Left Lung & Both (Ours) \\
    \midrule
   DSC & 0.67 & 0.28 & \textbf{0.83} \\
    \bottomrule
  \end{tabular}
  \caption{CLIP performance with separate prompts for left and right lungs, and combined prompts. }
  \label{tab:prompts}
\end{table}

Recent research indicates that CLIP performance can be enhanced through the utilization of visual prompting~\cite{yang2024fine, shtedritski2023does, yao2024cpt, subramanian2022reclip}. Visual prompting (VPT) involves the addition of markers like colorful boxes or circles directly onto an image, aiding in highlighting specific targets in image-language tasks. This technique directs the attention of Vision-Language Models towards desired targets while maintaining the global context. Inspired by this, we also used visual prompting to further evaluate using a separate set of prompts for the lungs. We evaluated three different visual prompts: red bounding box, gray reverse blur, and contour. In our case, we add visual markers on the original image around SAM-generated masks, and pass this set of images to CLIP, as discussed in supplementary~\cref{sec:vpt}. However, for medical datasets, VPT did not perform well. In contrast to such techniques,  our method, which involves employing a set of crops of the image according to SAM-generated masks, even while utilizing separate prompts for the left and right lung, still achieves a superior DSC of 0.65 as compared to other prompts as shown in~\cref{tab:vpt}.
\begin{table}
  \centering
  \begin{tabular}{@{}lcccc@{}}
    \toprule
   Prompt & Box & Reverse blur & Contour &  Crops (ours)\\
    \midrule
   DSC & 0.49 & 0.60 & 0.61 & \textbf{0.65}\\
    \bottomrule
  \end{tabular}
  \caption{Evaluation of visual prompting.}
  \label{tab:vpt}
\end{table}

These limitations have offered valuable insights into the failure cases. In the future, we aim to incorporate inference mechanisms to detect such failures and prevent their propagation to the subsequent steps in the pipeline. It will help mitigate the occurrence of such failures and improve performance further.

\subsection{Ablations}


\subsubsection{Different SAM models}
In this section, we assess whether employing different variants of SAM can improve the performance. We evaluated all three different versions: ViT-B (base), ViT-L (large), and ViT-H (huge). The results are presented in ~\cref{tab:sam_models}. Given its superior performance, we opt ViT-H version in our pipeline. Notably, as our approach is training/fine-tuning free and performs test time adaptation for zero-shot segmentation, integrating the ViT-H version imposes no extra training overhead.

\begin{table}[!hb]
\centering
\begin{tabular}{@{}lccc@{}}
\toprule
Dataset & ViT-B  & ViT-L & ViT-H  \\
\midrule
CC359~\cite{souza2018open} & 0.80  &0.89& \textbf{0.94}\\
X-ray~\cite{pandey2019} & 0.71 &  0.76 & \textbf{0.83}\\
HC18~\cite{Heuvel2018us} & 0.66 & 0.76 & \textbf{0.81}\\
\bottomrule
\end{tabular}

\caption{Ablation: Comparison of SAM's variant.}
\label{tab:sam_models}
\end{table}

\subsubsection{SaLIP vs SAM + CLIP}

To assess the performance enhancement brought by our proposed stacking approach of SaLIP, we compared it to SAM + CLIP. Unlike our framework,  SAM + CLIP utilizes $\mathbf{\textit{SAM}}_{\textbf{\textit{EM}}}$ and CLIP exclusively, with the CLIP-retrieved crop considered as the prediction. The results are presented in \cref{tab:sam_clip}. Our proposed approach leads to improvement.

\begin{table}[t]
\centering
\begin{tabular}{@{}lcc@{}}
\toprule
Dataset & SAM-CLIP  & SaLIP \\
\midrule
CC359~\cite{souza2018open} & 0.89 & \textbf{0.94}\\
X-ray~\cite{pandey2019} & 0.80 & \textbf{0.83}\\
HC18~\cite{Heuvel2018us} & 0.78  & \textbf{0.81}\\
\bottomrule
\end{tabular}
\caption{Ablation: Performance comparison between SAM-CLIP and SaLIP (ours).}
\label{tab:sam_clip}
\end{table}

\subsubsection{Area based filtering}
\label{area}
\begin{figure}[t]
  \centering
   \includegraphics[width=0.6\linewidth, height = 7cm]{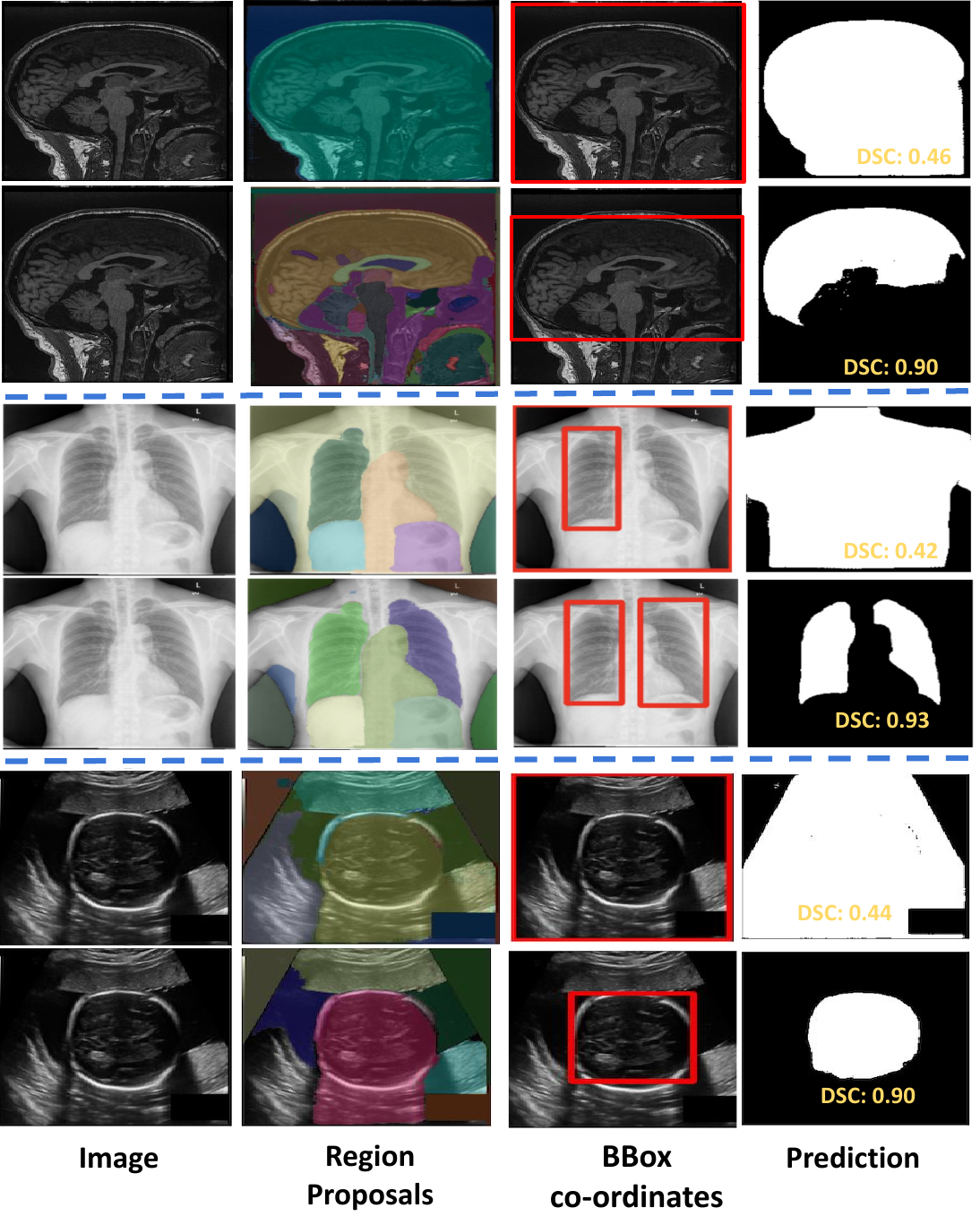}
   \caption{Ablation: the first row for each modality shows results without area filtering, and the second row illustrates the effects of area filtering (ours). }
   \label{fig:area}
\end{figure}

$\mathbf{\textit{SAM}}_{\textbf{\textit{EM}}}$ employs a grid-wise set of key points to generate masks for each part of the image. The resulting set of masks may include masks for the background or larger regions encompassing the region of interest (ROI). In such cases, CLIP can miss-classify background or larger region encompassing ROI as illustrated in ~\cref{fig:area}. To tackle this challenge, we implement area-based filtering, determining the optimal area threshold through a random hyperparameter search. To assess the effectiveness of this filtering approach, we compare its results with one obtained by passing all the masks generated by $\mathbf{\textit{SAM}}_{\textbf{\textit{EM}}}$ to CLIP without any filtering. The comparative results between area filtering and without area filtering are presented in ~\cref{tab:area}. Our area filtering approach shows an improvement of 3\% for brain segmentation \cite{souza2018open} and approximately 10\% for lung and fetal head segmentation \cite{pandey2019}.

\begin{table}[!hb]
\centering
\begin{tabular}{@{}lcc@{}}
\toprule
Dataset & No Filtering  & Filtering (Ours) \\
\midrule
CC359~\cite{souza2018open} & 0.91 & \textbf{0.94}\\
X-ray~\cite{pandey2019} & 0.75 & \textbf{0.83}\\
HC18~\cite{Heuvel2018us} & 0.71 & \textbf{0.81} \\
\bottomrule
\end{tabular}
\caption{Ablation: Impact of area-based filtering.}
\label{tab:area}
\end{table}





\label{clip_prompt_mode}


\section{Conclusion}
In this work, we propose a simple and effective unified framework SaLIP, that leverages the zero-shot segmentation and recognition capabilities of SAM and CLIP respectively. By harnessing both the segment everything and promptable segmentation modes from SAM, with CLIP acting as a bridge between them, we demonstrate effective zero-shot organ segmentation. Unlike other SAM-based segmentation methods, SaLIP is training/fine-tuning free and does not rely on domain expertise or annotated data for segmentation. It is fully adapted at test time without the need for pre-training or additional computational overhead. We employ SAM to segment each region within the image, then leverage CLIP to identify the region of interest by using visually descriptive prompts generated from GPT 3.5. Subsequently, we utilize the retrieved region of interest to prompt SAM for organ segmentation. We validate our framework across three diverse medical imaging datasets, demonstrating its robustness. Our work provides an in-depth exploration of SaLIP for zero-shot organ segmentation. In the future, we aim to expand this work to diverse medical imaging datasets and further improve SaLIP, by integrating an inference mechanism to avoid propagation of failures.

\section{Acknowledgment}
Dedicated to the memory of \textbf{\textit{Dr. Kevin McGuinness}}, whose unwavering support and passion for research will forever guide
us. 

This research is supported by Science Foundation Ireland under grant numbers 18/CRT/6183 (ML-LABS Centre for Research Training) and 12/RC/2289\_P2 (Insight Research Centre for Data Analytics). 



{
    \small
    \bibliographystyle{ieeenat_fullname}
    \bibliography{main}
}

\clearpage
\setcounter{page}{1}
\maketitlesupplementary
\label{supplementary}
\section{Qualitative Results}
\label{sec:qualitative_results}
\begin{figure}[!hb]
  \centering 
   \includegraphics[width=\linewidth]{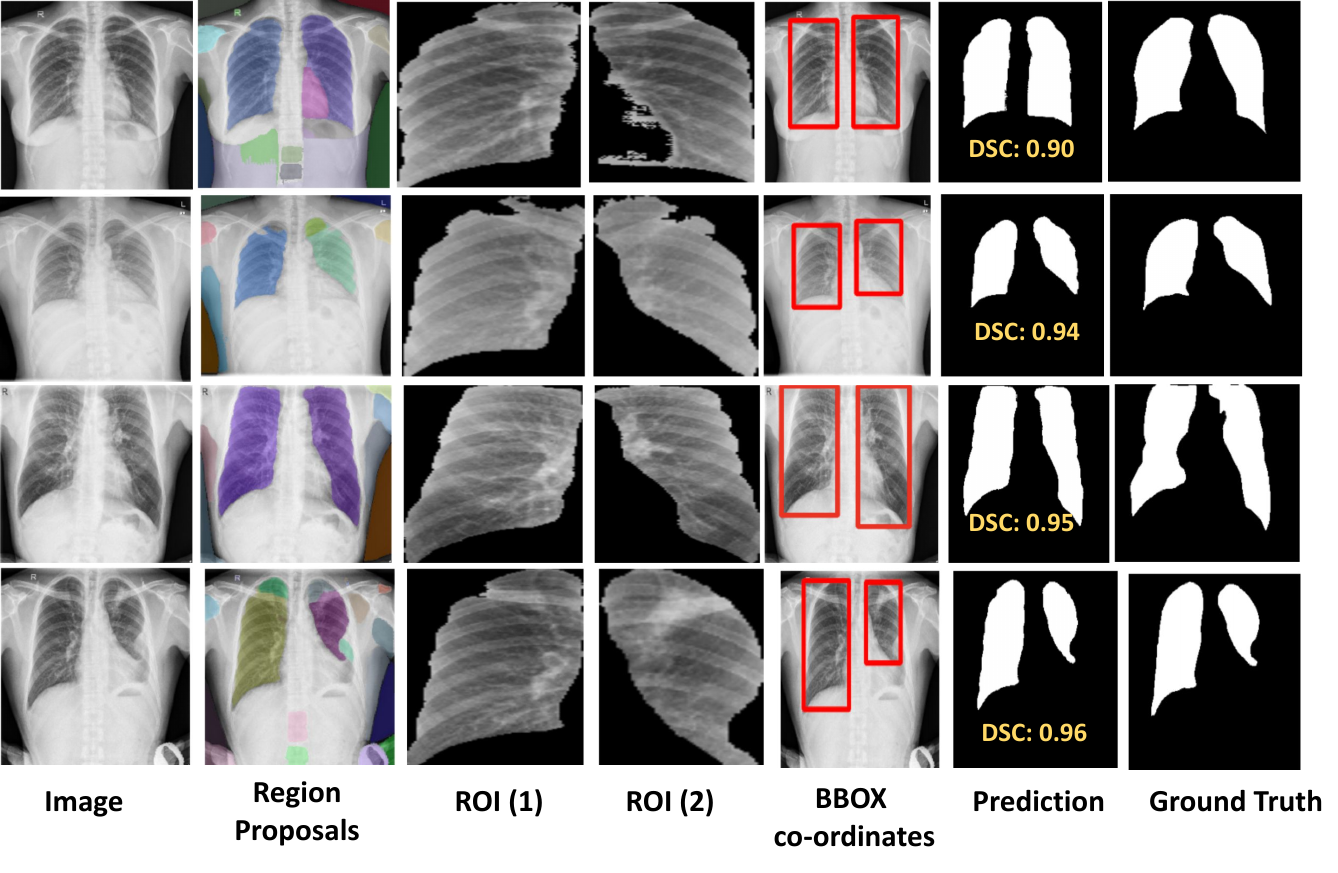}

   \caption{SaLIP qualitative results on X-ray labels and masks ~\cite{pandey2019} dataset.}
   \label{fig:lungs_preds}
\end{figure}

In this section, we comprehensively present the qualitative results of SaLIP to demonstrate its effectiveness. X-ray Masks and Labels ~\cite{pandey2019} dataset has two regions of interest i.e. left and right lungs. ~\cref{fig:lungs_preds} shows that SaLIP yields correct predictions for both lungs in a zero-shot manner without any domain expertise or prompt engineering.

\begin{figure}[!hb]
  \centering 
   \includegraphics[width=0.95\linewidth]{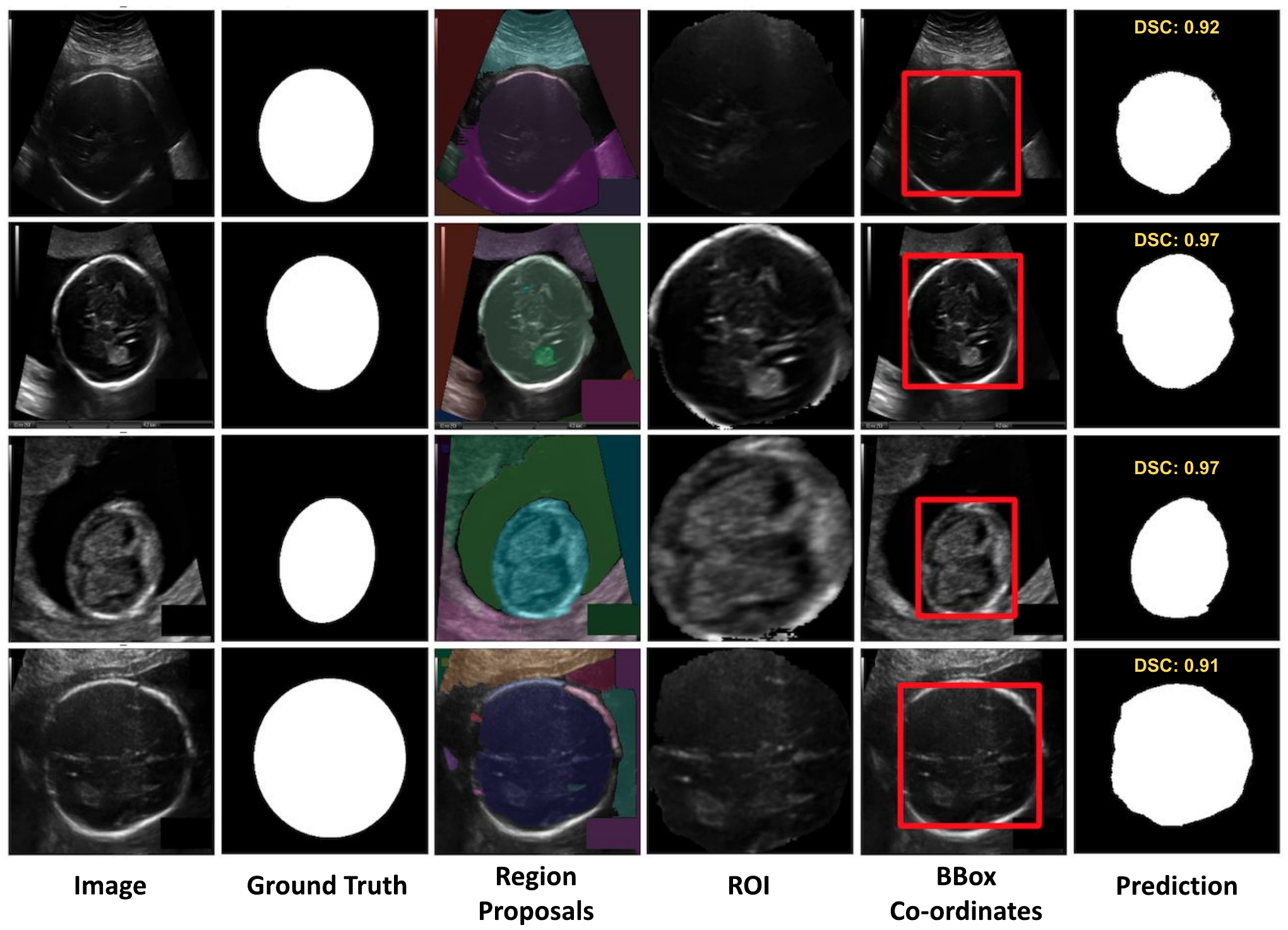}

   \caption{SaLIP Qualitative results on HC18 ~\cite{Heuvel2018us} dataset.}
   \label{fig:sam_success_cases}
\end{figure}

~\cref{fig:sam_success_cases} demonstrates that SaLIP accurately predicts fetal head positions with zero-shot learning. Despite the variation in fetal head size, SaLIP accurately segments them. 

\subsection{Limitations}
\label{sec:limitation_visulization}
\begin{figure}
  \centering 
   \includegraphics[width=\linewidth]{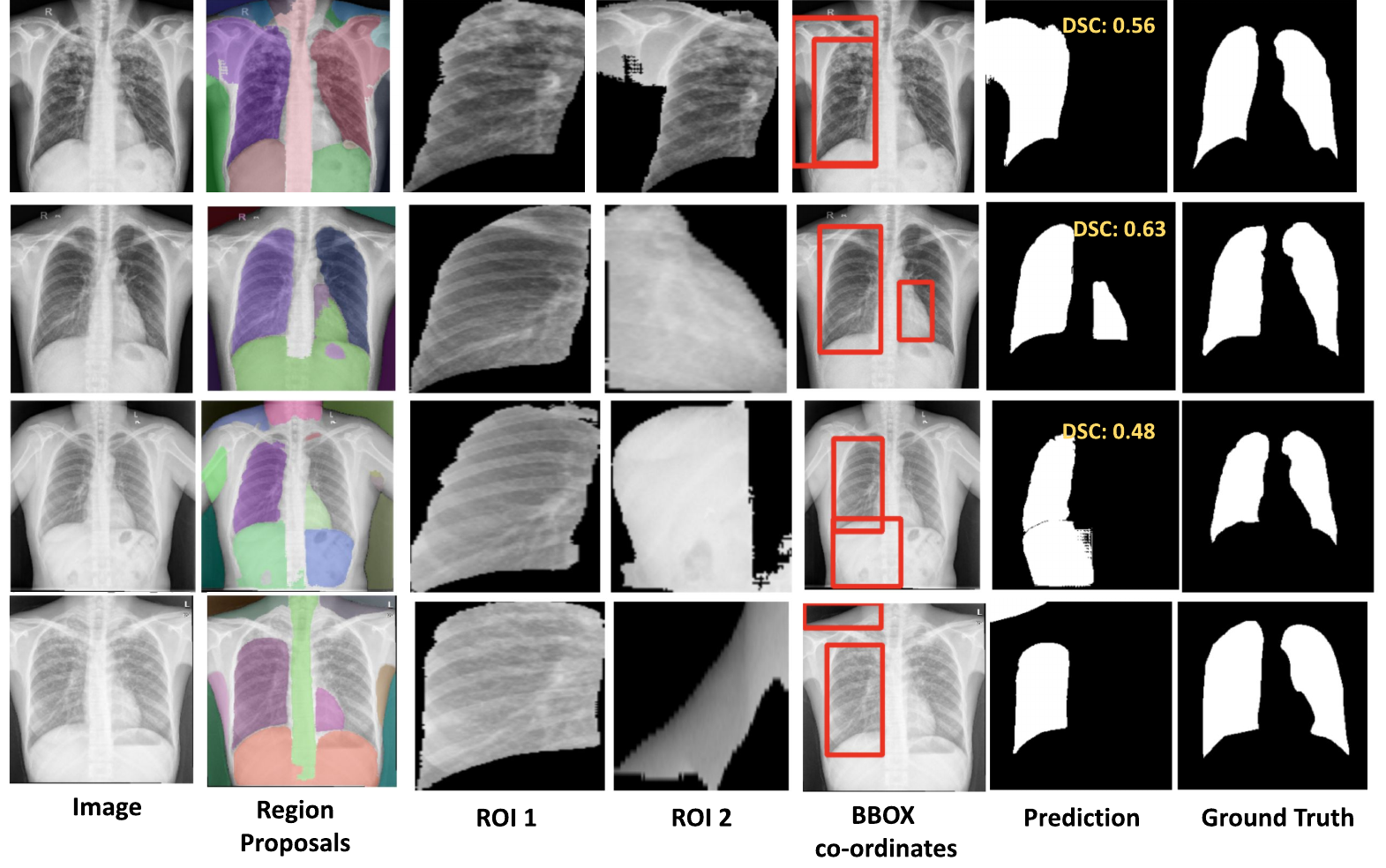}

   \caption{Visualization of failure cases on X-ray labels and masks dataset~\cite{pandey2019}. First two rows: SAM generates masks for both the left and right lung, but CLIP retrieves only the right lung mask correctly. While rows three and four show cases where SAM does not generate a mask for the left lung, eventually CLIP does not recognize the correct crop for the left lung in such cases. }
   \label{fig:lungs_failure}
\end{figure}

\begin{figure}
  \centering 
   \includegraphics[width=0.95\linewidth]{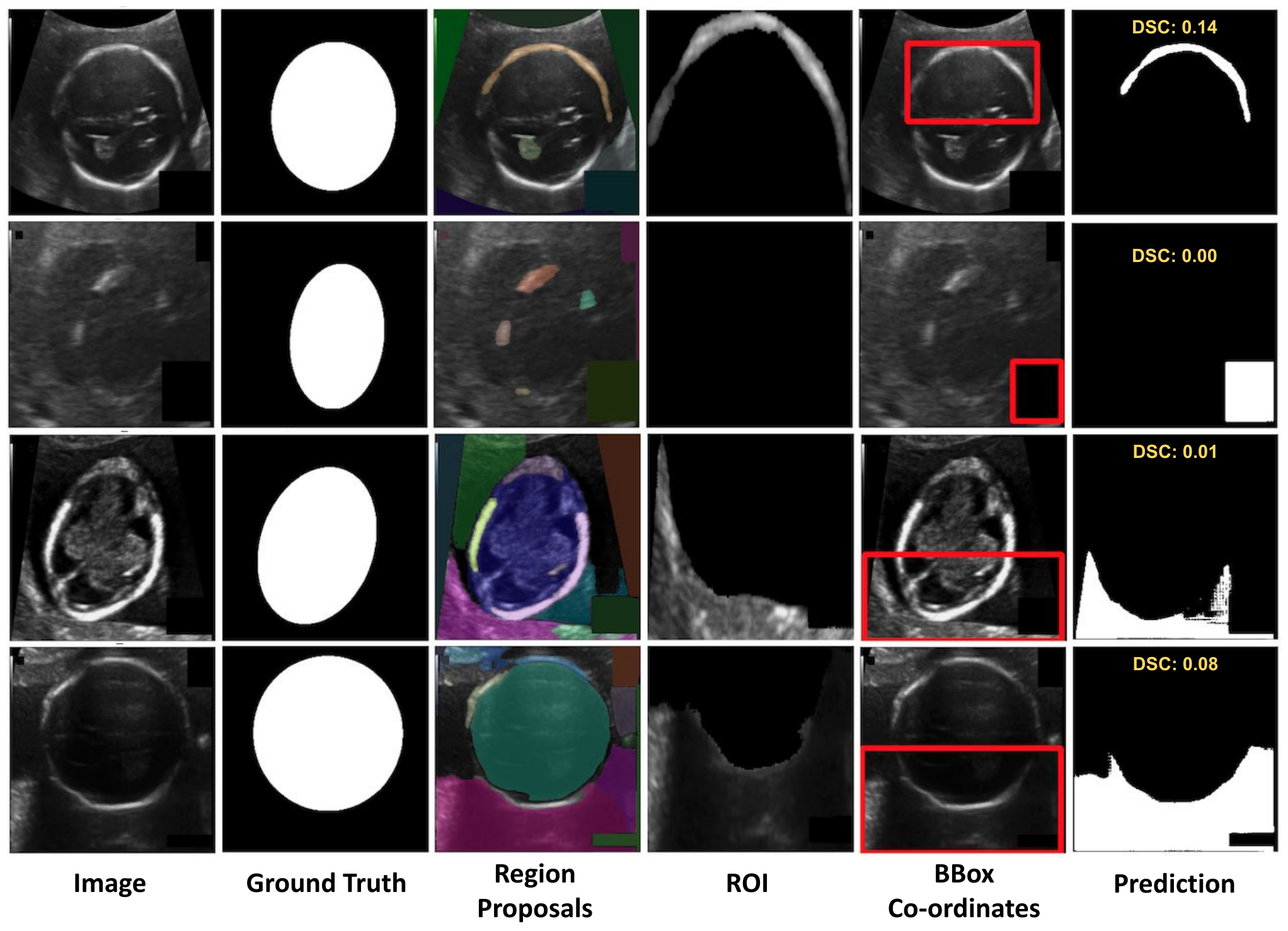}
   \caption{Visualization of failure cases for HC18~\cite{Heuvel2018us}. First two rows: $\mathbf{\textit{SAM}}_{\textbf{\textit{EM}}}$ does not generate a mask for the fetal head; therefore, CLIP fails to identify the correct crop in these instances. While row three and four show instances where $\mathbf{\textit{SAM}}_{\textbf{\textit{EM}}}$ successfully generates a mask for the fetal head, whereas CLIP does not retrieve the correct mask.}
   \label{fig:sam_failure_cases}
\end{figure}

In this section, we highlight the limitations of our method. We report the limitations both at SAM and CLIP levels as shown in ~\cref{fig:lungs_failure}. The first row shows the case multiple mask generations by SAM leads to the misclassification of ROIs by CLIP. Specifically, CLIP erroneously predicts the SAM region mask for the lungs, thereby impacting the final prediction. The second row shows the case where SAM correctly generates masks for both ROIs, however, CLIP misclassifies the crop for the left lung. The third and fourth rows illustrate scenarios where SAM fails to predict the ROI for the left lung, consequently leading to CLIP retrieving the incorrect crop region.
We also report the limitations in the fetal head dataset as shown in ~\cref{fig:sam_failure_cases}. The first and second rows show instances where $\mathbf{\textit{SAM}}_{\textbf{\textit{EM}}}$ fails to generate the relevant mask for the fetal head ROIs, which leads to incorrect ROI recognition by CLIP. The third and fourth rows showcase which $\mathbf{\textit{SAM}}_{\textbf{\textit{EM}}}$ generates multiple masks for fetal head ROIs, resulting in miss-identification by CLIP. Such misidentifications lead to inaccurate predictions for the fetal head, consequently lowering the DSC.


\section {Visual prompting}
\label{sec:vpt}
\begin{figure}[t]
  \centering
   \includegraphics[width=0.8\linewidth]{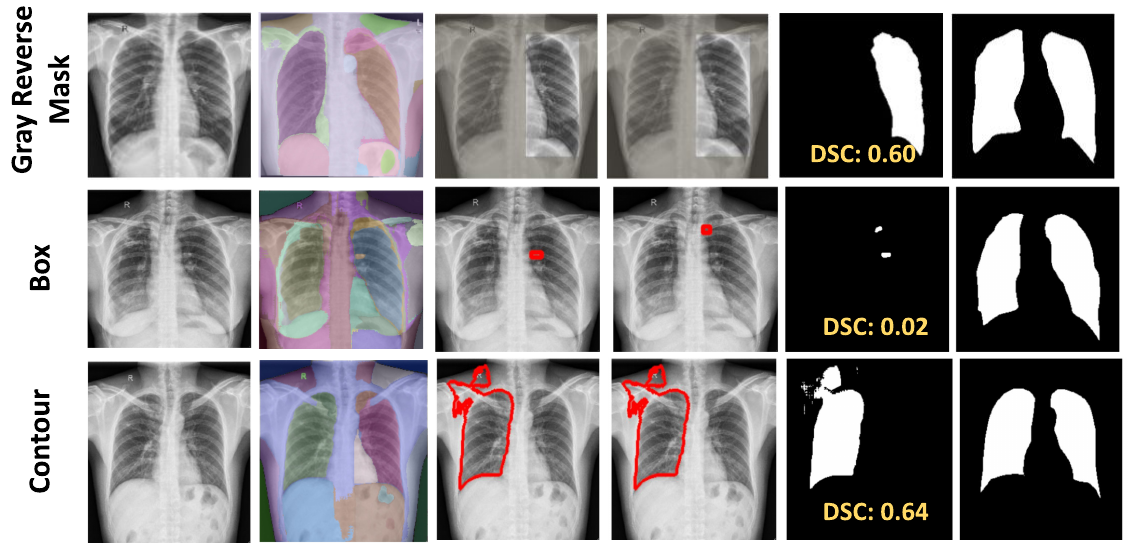}

   \caption{VPT results on X-ray labels and masks ~\cite{pandey2019} dataset: Although SAM successfully generates masks for both lungs, VPTs do not facilitate CLIP in accurately retrieving ROIs. }
   \label{fig:vpt}
\end{figure}

In this section, we provide details of different visual prompts around all the SAM's generated masks to improve lung segmentation. We applied three distinct visual prompts (VPTs) to all SAM-generated masks on the input image: a red bounding box marker, contour delineation, and a reverse gray box highlighting the masked area while blurring the rest of the image \cite{yang2024fine}. However, VPTs do not improve performance for medical imaging datasets as depicted in ~\cref{fig:vpt}. While in certain instances a single ROI is predicted correctly, the overall recognition is inconsistent across the dataset, indicating scalability issues.

\section{GPT-3.5 Prompts}
\label{sec:prompts}

\subsection{Lungs}
\label{sec:lungs}

In this study, we use a two-step prompting strategy to identify relevant characteristics that aid in the detection of crops having the desired organ, following the approach ~\cite{maniparambil2023enhancing}.

Example of the first prompt for the lung dataset:\\

\textit{I am creating a zero-shot lung recognition system in a chest x-ray. Write me 20 attributes to describe the lungs in the chest x-ray.}

Creating a zero-shot lung recognition system involves identifying and describing features that can uniquely identify and describe the lungs in a chest X-ray. Here are 20 attributes that could be relevant for such a task:\\

\textit{Lung Size: The overall size of the lungs, noting whether they are expanded, normal, or show signs of constriction or atelectasis.
Lung Shape: The general shape of the lungs, including any deviations from the typical butterfly shape.
Opacity: The degree of darkness observed in the lung fields, indicating the presence of fluids, masses, or other abnormalities.
Texture: The internal texture of the lung fields, whether it appears smooth, granular, or shows specific patterns such as reticular, nodular, or cystic formations.
Symmetry: The comparative symmetry between the right and left lungs, noting any significant differences.
Lung Borders: The clarity and integrity of the lung borders, particularly against the heart and diaphragm.
Hilum Position and Appearance: The location and appearance of the lung hila, which may be displaced or show signs of enlargement.
Pleural Line: The visibility and condition of the pleural line, including any thickening, plaques, or pneumothorax (air in the pleural space).
Costophrenic Angles: The sharpness and clarity of the costophrenic angles, noting any blunting which may suggest pleural effusion.
Vascular Markings: The visibility and pattern of vascular markings within the lung fields.
Air Bronchograms: The presence of air-filled bronchi (tubular structures) against a background of opaque lung, indicating consolidation.
Tracheal Position: The position of the trachea, noting if it is centered or displaced.
Diaphragm Appearance: The contour and position of the diaphragm, including any signs of elevation or abnormal curvature.
Cardiac Silhouette: The size and shape of the heart shadow as it relates to the lungs.
Mediastinal Width: The width of the mediastinum, noting any widening that may suggest mediastinal masses or lymphadenopathy.
Interstitial Markings: The presence of lines and markings that may indicate interstitial lung disease.
Lung Apex: The condition and appearance of the lung apex, looking for masses, fibrosis, or other abnormalities.
Presence of Calcifications: The presence and distribution of calcifications within the lung fields, which may suggest old infections or tumors.
Pleural Effusion: The presence of fluid in the pleural space, visible as an area of increased opacity.
Foreign Bodies or Devices: The presence of any foreign bodies, surgical clips, pacemaker leads, or other medical devices.
These attributes are critical for a detailed analysis and recognition of lung health and disease in chest X-rays. Each attribute can help differentiate between normal lung anatomy and various pathologies, aiding in the development of a comprehensive zero-shot lung recognition system.}

Example of the second prompt for the lung dataset which uses GPT-3.5's response for the first prompt as part of the prompt:

\textit{Now write me sentence about each of these attributes with "lung" in the sentences and return the answer as a Python dictionary.} \\

\textit{["The X-ray shows the shape of the lungs, whether they appear rounded or irregular.",
    "The X-ray indicates the relative size of the lungs within the chest cavity, providing insights into lung health.",
    "The X-ray helps assess the proper position of the lungs within the chest cavity.",
    "The X-ray reveals the density of the lung areas, indicating the presence of air or tissue within the lungs.",
    "The X-ray evaluates how clear the lung fields appear by assessing the transparency of the lungs.",
    "The X-ray shows the texture of the lungs, whether they appear smooth or have irregularities.",
    "The X-ray provides insights into the vascular pattern within the lungs, aiding in the assessment of lung circulation.",
    "The X-ray examines the borders of the lungs, assessing for normal anatomy or potential abnormalities.",
    "The X-ray checks the presence and integrity of lung fissures, the thin lines separating lung lobes.",
    "The X-ray evaluates structures at the lung hilum, such as lymph nodes and bronchi, for any abnormalities.",
    "The X-ray assesses the contour of the diaphragm where it meets the lungs, ensuring proper lung function.",
    "The X-ray checks the angles formed by the diaphragm and the chest wall, assessing for effusion or pneumothorax in the lungs.",
    "The X-ray examines the appearance and distribution of bronchi and blood vessels within the lungs.",
    "The X-ray detects fine linear or reticular opacities within the lung fields, indicating interstitial lung disease.",
    "The X-ray assesses the space between the lung and the chest wall for any effusion or pneumothorax.",
    "The X-ray evaluates the deviation of the trachea from its normal position, indicating potential mediastinal shift or mass effect on the lungs.",
    "The X-ray examines the appearance of ribs surrounding the lungs, checking for integrity and potential fractures.",
    "The X-ray shows the outline of the heart within the chest, which can influence the appearance of the lung fields.",
    "The X-ray assesses the structures within the mediastinum, providing insights into their impact on the adjacent lungs." ] }
\subsection{Separate prompts for left and right lung}
\label{sec:separate_prompts}
Example of first prompt for the lung dataset:\\

\textit{I am creating a zero shot right and left lung recognition system in a chest x-ray. Write me 1o attributes to describe the left and right lungs in the chest x-ray.}

Attributes of the Left Lung:

["Position: Situated on the left side of the thoracic cavity."
"Size: Slightly smaller than the right lung to accommodate the heart."
"Lobes: Consists of two lobes - the superior and inferior lobes, separated by the oblique fissure."
"Function: Main function is to exchange oxygen and carbon dioxide with the blood during respiration."
"Medical Significance: Annotated images showing the left lung are essential in medical imaging for diagnosing various respiratory conditions."
"Blood Circulation: Receives oxygenated blood from the pulmonary veins and sends deoxygenated blood to the pulmonary arteries."
"pH Regulation: Plays a role in regulating blood pH through the removal of carbon dioxide."
"Development: Undergoes morphological changes during development, including the formation of bronchial buds and bronchopulmonary segments."
"Susceptibility: Susceptible to infections such as pneumonia and tuberculosis, which can cause inflammation and compromise respiratory function."]

Attributes of the Right Lung:

["Position: Located on the right side of the thoracic cavity."
"Size: Slightly larger than the left lung due to the presence of the heart."
"Lobes: Divided into three lobes - the superior, middle, and inferior lobes, separated by the horizontal and oblique fissures."
"Function: Plays a crucial role in oxygenating the blood and removing carbon dioxide from the body."
"Medical Significance: Medical annotations on images showing the right lung aid in diagnosing diseases such as pneumonia, lung cancer, and tuberculosis."
"Blood Circulation: Receives deoxygenated blood from the systemic circulation via the pulmonary arteries and returns oxygenated blood to the heart through the pulmonary veins."
"Aspiration Risk: More prone to injury from aspiration due to its more vertical orientation of the main bronchus compared to the left lung."
"Diseases: Susceptible to diseases such as emphysema, bronchitis, and pulmonary embolism, which can impair respiratory function."
"Development: Undergoes branching morphogenesis during embryonic development to form the bronchial tree and alveoli."
"Gas Exchange: Has a larger capacity for gas exchange compared to the left lung due to its larger surface area."]

Example of the second prompt for the lung dataset which uses GPT-3.5's response for the first prompt as part of the prompt:\\

\textit{Now write me a sentence about each of these attributes with "left lung"  and "right lung" in the sentences of each respective class and return the answer as a Python dictionary.} \\

\{
    "left lung": [
        "A left lung is situated on the left side of the thoracic cavity.",
        "It is slightly smaller than the right lung to accommodate the heart.",
        "The left lung consists of two lobes: the superior and inferior lobes, which are separated by the oblique fissure.",
        "Its main function is to exchange oxygen and carbon dioxide with the blood during respiration.",
        "Annotated images showing the left lung are essential in medical imaging for diagnosing various respiratory conditions.",
        "The left lung receives oxygenated blood from the pulmonary veins and sends deoxygenated blood to the pulmonary arteries.",
        "In addition to its respiratory function, the left lung also plays a role in regulating blood pH through the removal of carbon dioxide.",
        "Certain diseases, such as pulmonary edema and pulmonary fibrosis, can affect the left lung's ability to function properly.",
        "During development, the left lung undergoes morphological changes, including the formation of bronchial buds and the establishment of bronchopulmonary segments.",
        "The left lung is susceptible to infections, such as pneumonia and tuberculosis, which can cause inflammation and compromise its respiratory function."
    ],
    \\
    "right lung": [
        "A right lung is located on the right side of the thoracic cavity.",
        "It is slightly larger than the left lung due to the presence of the heart.",
        "The right lung plays a crucial role in oxygenating the blood and removing carbon dioxide from the body.",
        "Medical annotations on images showing the right lung aid in diagnosing diseases such as pneumonia, lung cancer, and tuberculosis.",
        "The right lung is divided into three lobes: the superior, middle, and inferior lobes, separated by the horizontal and oblique fissures.",
        "It receives deoxygenated blood from the systemic circulation via the pulmonary arteries and returns oxygenated blood to the heart through the pulmonary veins.",
        "The right lung is more prone to injury from aspiration due to its more vertical orientation of the main bronchus compared to the left lung.",
        "Diseases affecting the right lung include emphysema, bronchitis, and pulmonary embolism, which can impair its respiratory function.",
        "During embryonic development, the right lung undergoes branching morphogenesis to form the bronchial tree and alveoli.",
        "The right lung has a larger capacity for gas exchange compared to the left lung due to its larger surface area."
    ]
\}

\subsection{Fetal Head }
Example of the first prompt for the fetal head  segmentation:\\

\textit{I am creating a fetal head recognition system from the transthalamic and transventricular ultrasound planes; write me 20 attributes to describe the fetal head.}\\

\textit{Head Circumference: Measurement of the circumference of the fetal head, indicating overall size. Biparietal Diameter (BPD): Distance between the two parietal bones, a key indicator of fetal head size and gestational age. Head Shape: Describing the overall shape of the fetal head, which can vary from round to more elongated. Skull Density: Density of the fetal skull bones, influencing ultrasound image contrast. Fontanelle Size: Size and appearance of the fontanelles, indicating the stage of skull ossification. Cerebral Ventricles: Presence and size of cerebral ventricles within the fetal brain. Cranial Bone Integrity: Assessment of the integrity of the cranial bones, looking for any fractures or abnormalities. Midline Structures: Identification of midline structures such as the falx cerebri and cavum septi pellucidi. Cerebellar Vermis: Presence and size of the cerebellar vermis, important for assessing brain development. Third Ventricle: Measurement and appearance of the third ventricle, indicative of normal brain development. Cerebral Hemispheres: Assessment of the size and symmetry of the fetal cerebral hemispheres. Occipitofrontal Diameter (OFD): Measurement from the occipital bone to the frontal bone, providing another indicator of fetal head size. Nuchal Fold Thickness: Thickness of the nuchal fold, which can indicate chromosomal abnormalities. Facial Features: Identification and characterization of facial features such as the nose, lips, and orbits. Fetal Neck: Assessment of the fetal neck for any abnormalities or masses. Eyes and Eye Orbit: Identification of the eyes and assessment of the eye orbits for normal development. Temporal Bone Anatomy: Examination of the temporal bones for any abnormalities or ossification issues. Brain Symmetry: Assessment of symmetry in brain structures, including the hemispheres and ventricles. Brain Parenchyma Texture: Evaluation of the texture and appearance of the fetal brain parenchyma. Fetal Presentation: Determination of the fetal presentation (cephalic, breech, transverse), which can impact the segmentation process.}\\

Example of the second prompt for the fetal head dataset which uses GPT-3.5's response for the first prompt as part of the prompt:\\

\textit{Now write me sentences about each prompt of these attributes with "fetal head" in the sentences, and return the answer as a python dictionary.}

The response of the second prompt for fetal head as an example:\\

\textit{["Fetal head circumference is a crucial metric measured to assess overall growth and development.", "Biparietal diameter (BPD) provides insight into fetal head size and gestational age.", "The fetal head's shape varies, ranging from round to more elongated forms.", "Skull density contributes to the clarity and contrast of structures within the fetal head.", "Fontanelle size offers information about the fetal head's ossification stages and cranial development.", "Cerebral ventricles within the fetal head are assessed to ensure normalcy and appropriate size.", "Cranial bone integrity is essential for the protection and development of the fetal head.", "Midline structures, such as the falx cerebri, contribute to the overall architecture of the fetal head.", "The cerebellar vermis, located within the fetal head, influences motor coordination and balance.", "The third ventricle within the fetal head is examined to evaluate central nervous system development.", "Symmetry of the cerebral hemispheres is crucial for the overall morphology of the fetal head.", "Occipitofrontal diameter (OFD) measurement aids in assessing the proportionality of the fetal head.", "Nuchal fold thickness measurement provides insights into potential abnormalities affecting the fetal head.", "Facial features within the fetal head region contribute to the overall assessment of anatomical development.", "Evaluation of the fetal neck ensures proper alignment and structural integrity of the fetal head and neck.", "Eyes and eye orbits within the fetal head region are assessed to evaluate ocular development.", "Temporal bone anatomy within the fetal head is evaluated to detect any abnormalities or ossification issues.", "Brain symmetry within the fetal head is assessed to ensure proper development and alignment of structures.", "Texture and appearance of the brain parenchyma provide insights into overall health and development within the fetal head.", "Fetal presentation, whether cephalic, breech, or transverse, influences the positioning and segmentation of the fetal head during imaging."
]}







%
%
%

\end{document}